\documentclass{article}
\newcommand\vv[1]{{\overline{#1}}}

\newtheorem{theorem}{Theorem}

\newlength{\proofbox}
\newenvironment{proof}{\begin{description}\item[Proof:]}{
\hspace*{\fill}\rule{\proofbox}{\proofbox}
\end{description}}
\advance\hoffset by -.7in
\title{{\bf A Hierarchical Situation Calculus}}
\author{
David A. Plaisted\thanks{This research was partially supported by the
National Science Foundation under grant CCR-9972118.}\\
Department of Computer Science\\
UNC Chapel Hill\\
Chapel Hill, NC 27599-3175\\
Phone: (919) 967-9238\\
Fax: (919) 962-1799 \\
{\small Email: plaisted@cs.unc.edu}}
\begin{document}
\bibliographystyle{alpha}
\maketitle
\begin{abstract}
A situation calculus is presented that provides a solution to the
frame problem for hierarchical situations, that is, situations that
have a modular structure in which parts of the situation behave in a
relatively independent manner.  This situation calculus is given in a
relational, functional, and modal logic form.  Each form permits both
a single level hierarchy or a multiple level hierarchy, giving six
versions of the formalism in all, and a number of sub-versions of
these.  For multiple level hierarchies, it is possible to give
equations between parts of the situation to impose additional
structure on the problem.  This approach is compared to others in the
literature.
\end{abstract}
\section*{Keywords}
Situation calculus, frame problem, modal logic

\section{Introduction}
The situation calculus formalism permits reasoning about properties of
situations that result from a given situation by sequences of actions
\cite{McHa:69}.  A problem with this formalism is the necessity to
include a large number of {\em frame axioms} that express the fact
that actions do not influence many properties of a situation (for
example, if you walk in a room, the color of the wall typically does
not change).  Non-monotonic logic is one approach for handling such
frame axioms, by assuming that properties stay the same unless they
can be proved to change, but this logic has additional complexity
compared to classical logic.  Reiter \cite{Reiter:91} proposed an
approach to the frame problem in first-order logic that avoids the
need to specify all of the frame axioms.  In this paper another
approach to the frame problem based on {\em aspects} is presented.  An
aspect is essentially a relationship between a situation and one of
its component parts.  Each action and fluent (property of a situation)
has an aspect, indicating which component of the situation the aspect
or fluent refers to, and if the aspects satisfy a certain specified
relation, it follows that the action does not influence the fluent.
For example, if an action and fluent have different aspects, this
can indicate that the action influences a different component of the
situation than the fluent refers to.
Using aspects to specify frame axioms
can reduce their complexity, because many actions and fluents
can have the same aspect.  In addition to the use of aspects for
specifying frame axioms,
a number of approaches for deriving aspects
of actions and fluents from
more fundamental properties of a situation are also
given.

\section{Aspects}

All methods for formalizing aspects considered here
have the following general features: Actions modify situations, which
are denoted by the variables $s$, $t$, and $u$.  A situation is
assumed to be composed of relatively independent parts and each part
is called an {\em aspect} of the situation.  Thus if a building has
four rooms, each room could be considered an aspect of the building,
because many actions will only influence one room of the building.
In the traditional situation calculus, a situation $s$ represents
the entire world.  In the aspect formalism, a situation $s$ can represent
just a portion of the world, for example, a room of a building.
If $\alpha$ is an aspect, then $R_{\alpha}(s,t)$ indicates that
situation $t$ is an $\alpha$ aspect of situation $s$.  If $t$ is
unique, it may be denoted by $f_{\alpha}(s)$.
The symbol $a$ denotes an action, and $a(s)$ denotes the situation
that results when action $a$ is performed in situation $s$.  This is
an abbreviation for $do(a,s)$, for example, $do(drop(x),s)$.  Also,
$a$ may have preconditions, but it is not necessary to consider them
here.  A {\em fluent} is a relation or predicate that can be true or
false in a situation.  The symbols $p$ and $q$ denote fluents.  The
notation $p(s)$ indicates that fluent $p$ is true in situation $s$.
This is an abbreviation for $holds(p,s)$, for example,
$holds(at(x,y),s)$.  In the aspect formalism, some fluents may be
properties of a portion of the world and others may not.  If fluent
$p$ is not a property of the situation $s$, because $p$ refers to
something that is outside of the portion of the world modeled by $s$,
then $p(s)$ is undefined.  (For our purposes, $p(s)$ can have an
arbitrary truth value in this case.)  Similary, if an action $a$ does
not refer to the portion of the world modeled by $s$, then
$a(s)$ is undefined.
Actions and fluents can have aspects,
indicating which parts of the situation they influence.  For example,
if an action $a$ has as aspect a room of a building, this can indicate
that the action can only influence properties of that room.  If a
fluent has as aspect a room of a building, this can indicate that the
fluent is only a property of that room.  Thus if action $a$ and fluent
$p$ have different aspects, then performing $a$ will not affect $p$.
The symbols $\alpha_i$ and $\beta_i$ denote aspects of fluents and
actions, respectively, and are drawn from universes ${\cal A}_1$ and
${\cal A}_2$ of objects, respectively.  Thus $\alpha_i \in {\cal A}_1$
and $\beta_i \in {\cal A}_2$.
Often ${\cal A}_1 = {\cal A}_2$.  For
a fluent $p$, $p:\alpha$ indicates that $p$ has aspect $\alpha$.  For
an action $a$, $a:\beta$ indicates that action $a$ has aspect $\beta$.
Theoretically, a fluent or action could have more than one aspect, but
the following discussion assumes that aspects of fluents and actions
are unique.  However, there can be many different actions and fluents
that have the same aspect.  Therefore the number of aspects can be
much smaller than the number of fluents and actions.

For each problem domain and each formalism, it is necessary to give
axioms specifying the aspects of fluents and actions.
Let $F[a,p]$
be the {\em frame axiom}
\begin{equation}
\forall s (p(s) \equiv p(a(s)))
\end{equation}
indicating that action $a$ does not affect fluent $p$.  Let $d(\alpha,\beta)$
be a {\em non-interference} property, where $d$ is a predicate symbol;
this property will be specified differently
for each problem domain.
Intuitively, $d(\alpha,\beta)$ means that
actions of aspect $\beta$ do not influence fluents of aspect $\alpha$.
A typical non-interference specification would be
$(\alpha \neq \beta) \supset d(\alpha,\beta)$.  For example, an action
$a$ and a fluent $p$ that correspond to distinct rooms of a building
would not interact.
It is also possible to let 
aspects of fluents be elements of some set ${\cal F}$  
and let aspects of actions be sets of elements from ${\cal F}$
Thus 
${\cal A}_1 = {\cal F}$ and
${\cal A}_2 = 2^{\cal F}$.
Then the non-interference specification could be
$\alpha \not\in \beta \supset d(\alpha,\beta)$.
For example, an action $a$ might influence two rooms $r_1$ and
$r_2$ of a building;  then its aspect might be the set $\{r_1,r_2\}$.
A fluent $p$ that is only a property of room $r_3$ might have aspect
$r_3$.  Because $r_3 \not\in \{r_1,r_2\}$, performing action $a$ would
not influence fluent $p$.
Another example is a telephone call between locations $x$ and $y$
that can only influence fluents having aspect $x$ or $y$.
Because the number of aspects can be much smaller than the number of
fluents and actions, the axiomatization of $d$ can be small relative
to the axiomatization of the entire domain.
For each formalism it is necessary to show the {\em non-interference axiom}
\begin{equation}
p:\alpha \wedge a : \beta \wedge d(\alpha,\beta) \supset F[a,p]
\label{intro.2}
\end{equation}
for all $p, a, \alpha, \beta$.  If $p:\alpha$ and $a : \beta$ and
$d(\alpha,\beta)$, then $p$ and $a$ are said to be {\em disjoint}, otherwise
they {\em intersect}.
For each version of the formalism, it is only
necessary to give frame axioms for $a$ and $p$ that intersect.
This can considerably reduce the effort necessary to give frame axioms.

\subsection{Sequences of Aspects}
In formalisms based on sequences, actions and fluents have {\em
sequences} of aspects instead of single aspects.  The notation
$\vv{\alpha}$ denotes a sequence of aspects, that is, $\alpha_1,
\dots, \alpha_n$.  The sequences of aspects may have different lengths
for different actions and fluents.  For this formalism, one has
$d(\vv{\alpha},\vv{\beta})$ for the non-interference property.  The
non-interference axiom becomes
\begin{equation}
p:\vv{\alpha} \wedge a : \vv{\beta} \wedge d(\vv{\alpha},\vv{\beta})
\supset F[a,p] \label{seq.intro.2}
\end{equation}
for all $p, a, \vv{\alpha}, \vv{\beta}$.  If $p:\vv{\alpha}$ and $a :
\vv{\beta}$ and $d(\vv{\alpha},\vv{\beta})$, then $p$ and $a$ are said
to be {\em disjoint}, otherwise $p$ and $a$ {\em intersect}, as
before.  A typical axiom would be $\exists i(\alpha_i \neq \beta_i)
\supset d(\alpha,\beta)$.  Intuitively, sequences of aspects represent
more than one level hierarchy.  For example, an aspect of a country
could be a state and an aspect of a state could be a city in the
state.  Some actions might influence a whole country, some might
influence just one state, and some might influence just one city in
one state.  These would correspond to different length sequences of
aspects.  Another example is binary trees; a sequence of 0 and 1
specifies a subtree of a binary tree, with 0 signifying left subtree
and 1 signifying right subtree, for example.  So an empty sequence
refers to the whole tree, the sequence (1,1) refers to the right
subtree of the right subtree, et cetera.  If an action $a$ has aspect
(1,1) and a fluent $p$ has aspect (0,1,0) then the action can only
influence the right subtree of the right subtree, while the fluent is
a property of a different subtree.  Thus one would have $F[a,p]$.
If the action has aspect (1,1) and the fluent has aspect (1,1,0) then
the action can influence the fluent.  Longer sequences indicate
smaller influences of an action and fewer dependencies of a fluent.

\subsection{Assuming the Non-Interference Axiom}
The preceding discussion, both for aspects and sequences of aspects,
is concerned with justifying the non-interference axiom from more
basic properties.  A simpler approach would be to {\em assert} the
non-interference axiom for each problem domain; then it is only
necessary to assign aspects to fluents and actions and give axioms
specifying $d$.  Even this approach can greatly simplify the frame
axioms.  Suppose there are $m$ fluents $p_1, \dots, p_m$ having aspect
$\alpha$ and $n$ actions $a_1, \dots, a_n$ having aspect $\beta$.
Then from the non-interference axiom, the specification of $d$, and
the assertions $p_i : \alpha$, $1 \leq i \leq m$ and $a_j : \beta$, $1
\leq j \leq n$ it is possible to deduce the $mn$ frame axioms
\begin{equation}
F[a_j,p_i], 1 \leq i \leq m, 1 \leq j \leq n
\end{equation}
This approach requires only $m + n + 2$ small axioms, assuming that the
specification of $d$ requires only one small axiom, whereas directly
specifying the $mn$ frame axioms would require $mn$ axioms.  Of
course, there could be other actions and fluents having other aspects
$\beta'$ and $\alpha'$.  For each pair $(\alpha,\beta)$ of aspects
such that $d(\alpha,\beta)$, one obtains $mn$ frame axioms, where
there are $m$ fluents of aspect $\alpha$ and $n$ actions of aspect
$\beta$.  This approach essentially classifies actions and fluents
and uses this classification to formalize general frame axioms having
many specific frame axioms as consequences.  This method alone gives
an economical solution to the frame problem in some cases.

However, the question remains as to how the non-interference axiom is
justified.  It is profitable to consider various semantics and
axiomatizations from which the non-interference axiom is derivable.
This helps to gain insight into this axiom, as well as providing guidance
concerning when the non-interference property should be used and how
it can be formalized.

A number of approaches to axiomatizing the non-interference property
are given.  In many cases, more than one
formalism is applicable.  The choice between sequential and
non-sequential formalisms can be dictated largely by the problem
structure, whether the domain has a single level or multi-level
hierarchy.  The choice between existential and universal versions of
the relational formalisms is less clear.  The functional formalisms
appear simpler than the relational ones, and probably are preferable
when they apply.  They also may permit more equational reasoning.  It
would also be possible to have mixed universal and existential
relational formalisms.  The modal formalisms are equivalent to
the relational ones, and the choice between them may be a matter of
taste.

\subsection{Examples}
An example, the blocks world example, will illustrate some of the
features of the aspect formalism, even though this example is not
particularly well suited to aspects.  There are two fluents,
$on(x,y)$ (block $x$ is on block $y$) and $clear(x)$ (nothing is on
$x$).  The domain consists of blocks and the floor.  The action is
$move(x,y)$, move block $x$ on block $y$, with preconditions that both
$x$ and $y$ be clear.  Let the aspect axioms be $on(x,y) : y$,
$clear(x): x$, $on(x,z) \supset move(x,y): \{y,z\}$, and $\neg
(\exists z) on(x,z) \supset move(x,y): \{y\}$.  For simplicity,
assume every block is on something.
In general, the aspect
of a block is the block it rests on (or the floor), and moving a block
$x$ has as aspect both its old and new locations.  Let the
disjointness predicate $d$ be axiomatized by $\alpha \not\in \beta
\supset d(\alpha,\beta)$.

The general non-interference axiom, axiom \ref{intro.2}, has the
following instance:
\begin{equation}
clear(w):\alpha \wedge move(x,y) : \beta \wedge \alpha \not\in \beta
\wedge holds(clear(w),s) \supset holds(clear(w),do(move(x,y),s))
\end{equation}
From this instance
and the axioms $clear(w) : w$, $on(x,z) \supset move(x,y):\{z,y\}$, and $w
\neq y \wedge w \neq z \supset w \not\in \{z,y\}$ the following frame axiom
follows.
\begin{equation}
w \neq y \wedge w \neq z \wedge on(x,z) \wedge holds(clear(w),s)
\supset holds(clear(w),do(move(x,y),s))
\end{equation}
Note that the condition $w \neq z$ of the frame axiom is not
necessary, so this frame axiom is weaker than necessary.
For problem domains with more of a hierarchical structure, frame axioms
derived from aspects should generally be as strong as possible.
Another instance of the non-interference axiom is the following:
\begin{equation}
on(v,w):\alpha \wedge move(x,y) : \beta \wedge \alpha \not\in \beta
\wedge holds(on(v,w),s) \supset holds(on(v,w),do(move(x,y),s))
\end{equation}
By similar reasoning, the following frame axiom follows from this instance:
\begin{equation}
w \neq y \wedge w \neq z \wedge on(x,z) \wedge holds(on(v,w),s)
\supset holds(on(v,w),do(move(x,y),s))
\end{equation}
The condition $w \neq y$ is not necessary for this frame axiom.

The blocks world example can be made more hierarchical by assuming
that there are a number of rooms with blocks in each room.  An action
that only involves the blocks in one room, and does not move them to
another room, will not influence any other rooms.  An action that
moves a block from one room to another can influence both rooms.  For
the multi-room example, the aspect of a block can be a sequence
$(\alpha_1,\alpha_2)$ where $\alpha_1$ gives the room the block is in
and $\alpha_2$ gives the block it rests on.  The aspects of fluents
and actions are obtained from the blocks they refer to.  Then $d$ can
be defined so that an action and fluent referring to different rooms,
are disjoint.

As another example illustrating sequences of aspects on a hierarchical
domain, consider actions that change some of the pixels on
a display and fluents that describe properties of a subset of the pixels.
Let ${\cal A}_1$ and ${\cal A}_2$ be the set of subsets of the pixels.
For each action $a_j$ suppose that $a_j : \beta$ where $\beta$ is the
set of pixels modified by the action, and for each fluent $p_i$ suppose
that $p_i : \alpha$ where $p_j$ is a property of the set $\alpha$ of pixels.
Then from the non-interference axiom one obtains the following:
\begin{equation}
p_i : \alpha \wedge a_j : \beta \wedge \alpha \cap \beta = \phi
\supset F[a_j,p_i].
\end{equation}
Now suppose that in addition there are actions that modify the memory
of a computer but not the display.  These can be given a ``memory''
aspect and the display actions and fluents can be given a ``display''
aspect to indicate that they do not interfere.  If there are other
objects in the room (such as a window and door) that are independent
of the computer then actions and fluents that refer to them can be
given ``window'' and ``door'' aspects, and all the computer actions
and fluents can be given a ``computer'' aspect, to indicate that they
do not interact.  Some actions may influence the whole room.  In this
way one
obtains a sequence of aspects; for display actions and fluents the
sequence is of the form $(computer, display, S)$ where $S$ is a set of
pixels.  For memory actions and fluents the sequence is $(computer,
memory, T)$ where $T$ is the set of memory cells affected by the
action.  Actions that influence the window and door would have
sequences of the form $(window)$ and $(door)$, respectively.  Actions
that influence the whole room, such as a meteorite hitting,
would have an empty sequence.  Let the
disjointness predicate $d$ be axiomatized by $d((\alpha_1, \dots,
\alpha_m), (\beta_1, \dots, \beta_n))$ if for some $i$, $\alpha_i \neq
\beta_i$.  Then if $\alpha_1, \dots, \alpha_m$ is the sequence of
aspects for a fluent $p$ and $\beta_1, \dots, \beta_n$ is the sequence
of aspects for an action $a$, the frame axiom $F[a,p]$ follows from
the non-interference axiom if for some $i$, $\alpha_i \neq \beta_i$.
Thus actions on the display do not influence the memory or the door or
the window, actions on the door or window do not influence the display
or the memory, but actions on the whole room may influence anything.

\subsection{Comparison to Other Approaches}
The most relevant comparable approach to the frame problem is that of
Reiter \cite{Reiter:91}, based on the work of Haas \cite{Haas:87},
Pednault \cite{Pednault:89}, Schubert \cite{schubert90monotonic}
and Davis \cite{Davis:02}.  Reiter essentially gives an axiom for each
fluent $p$ that says, For all actions $a$, if $a$ is possible in
situation $s$, then $p(a(s))$ holds if either (1) $a$ is an action
that makes $p$ true in situation $s$, or (2) $p(s)$ is already true
and $a$ is not an action that can make $p$ false in situation $s$.
One such axiom is needed for each fluent, so the number of such axioms
is equal to the number of fluents.  This approach avoids the frame
problem by focusing on actions that {\em change} fluents rather than
actions that do not change fluents.  Thus it is not necessary to give
all the frame axioms $F[a,p]$.

In particular, Reiter gives the following {\em successor state axiom}
for each fluent $R$:
\begin{equation}
Poss(a,s) \supset [R(do(a,s)) \equiv \gamma_R^+(a,s)\vee
R(s) \wedge \neg \gamma_R^-(a,s)]
\end{equation}
where $Poss(a,s)$ expresses that action $a$ is possible in situation $s$,
$\gamma_R^+(a,s)$ is true if action $a$ can make $R$ become true and
$\gamma_R^-(a,s)$ is true if action $a$ can make $R$ become false,
informally speaking.  Thus $R(do(a,s))$ is true if $a$ caused $R$ to
become true or if $R(s)$ was already true and $a$ did not cause $R$ to
become false.

A problem with Reiter's approach is that the successor state axiom for
$p$ can be very complicated, if many actions influence $p$; of course,
this complexity appears to be unavoidable in any approach.  Another
problem is that it is necessary to consider {\em all} actions.
Perhaps there are many actions that can influence $p$ but only a small
subset of them is necessary for a planning problem.  In fact,
knowledge concerning all actions that influence a fluent may not even
be available.  Reiter's approach requires that all of the actions be
mentioned, whether they will be used or not.  The traditional
formalization of the frame axioms avoids this problem.  The aspect
approach also avoids this problem.  But
it is straightforward to modify Reiter's approach
to handle unknown actions; it is only necessary to quantify the
successor state axiom over all $a$ in some set $S$ of actions that is
of interest, assuming that $S$ is sufficient for planning or reasoning
purposes.  It may also be true that for some action $a$ of interest one
does not know its effect on a fluent $R$.  In this case, the
corresponding predicate $\gamma_R^+(a,s)$ or $\gamma_R^-(a,s)$ need
not be axiomatized, if the remaining axioms are sufficient for
planning or reasoning purposes.

In Reiter's approach, to show that $R(a(s))$, assuming $R(s)$, it is
necessary to demonstrate $\neg \gamma_R^-(a,s)$.  This involves testing
$a$ against other actions, one by one, and verifying that $a$ is distinct
from all actions that can make $R$ false.  In the traditional approach,
the frame axiom $R(s) \supset R(a(s))$ serves the purpose much more
simply.  Proofs in the aspect based system are also simple and
intuitive.
Therefore proofs are more complex in Reiter's system than in
the traditional system and in the aspect based system.  Such complex
proofs may be more difficult for an automatic reasoning system to find.
The successor
state axiom in Reiter's system is compact, but unintuitive.  One would
presumably justify it by appealing to the traditional frame axioms
$F[a,p]$.  Therefore, for the sake of correctness, it may be preferable
to use the traditional frame axioms or the aspect
formalism.  Similarly, a proof from the
traditional frame axioms $F[a,p]$ or from axioms for aspects
may be more convincing to a user than
a proof from Reiter's successor state axiom.  Of course, it is possible
to obtain a proof using the successor state axiom and then modify it
to obtain a proof from the traditional frame axioms.

Reiter's approach has an extra cost, namely, the necessity to reason about
equality between actions.  The aspect formalism also has a cost, namely,
the need to reason about aspects.  Each approach may complicate the
reasoning process in a different way.  Reiter's definition of equality
between actions also means that equality on actions
is defined in a non-extensional way.

Another problem with Reiter's approach, noted in Scherl and Levesque
\cite{scherl93frame}, is that it can be difficult to incorporate {\em
constraints} between fluents, for example, when one fluent implies
another.  The successor state axiom essentially implies that the only
way a fluent can become true is if an action makes it true.  Lin and
Reiter \cite{lin94state} discuss how to overcome this by modifying the
successor state axiom.  For the aspect approach, constraints are not a
problem.

An advantage of Reiter's formalism is its conciseness.  The aspect
based formalism can also be concise, if for all but a small number of
action-fluent pairs $(a,p)$, $a$ and $p$ are disjoint, that is, $p :
\alpha \wedge a : \beta \wedge d(\alpha,\beta)$.  In such domains, the
aspect formalism may be preferable.  However, in other domains, the
conciseness of Reiter's formalism may be preferable.  In fact, the
aspect based system is concise if the number of pairs $(a,p)$ such
that $a$ and $p$
intersect is not much larger than the number of pairs $(a,p)$ such
that $F[a,p]$ is false, because the latter will contribute to the
complexity of either formalism, as well as to that of the traditional
formalism.

One can combine Reiter's approach with
aspects by using the successor state axiom {\em and} the
non-interference axiom.  When the non-interference axiom can be used
to demonstrate $F[a,p]$, it may provide simpler proofs than the
successor state axiom.  When $F[a,p]$ either is false or cannot be
shown using the non-interference axiom, the successor state axiom may
be used.
For this combined formalism it is also necessary to axiomatize
$d$, include the non-interference axiom (or other assertions from
which it can be derived),
and specify the aspects of actions and fluents.

A prior aspect formalism for the situation calculus
was presented in \cite{PlZh:97}.  This
formalism also treated hierarchical situations and defined aspects
of situations.  However, the axiomatization and semantics were
different.  The relation between a predicate $p$ of situation $s$
and a predicate $q$ on an aspect of $s$ was not considered.  Also,
there was no 
consideration of a situation variable as representing only one
component of a situation.

There have been many other approaches to the frame problem, but none
directly related to the aspect formalism.  Turner
\cite{representing97turner} shows how to represent commonsense
knowledge using default logic in the situation calculus.  He is
concerned not only with the truth of fluents but on causality
relations between them.  Pirri and Reiter \cite{pirri99some} formalize
situations as sequences of actions and discuss soundness and
completeness issues for planning with deterministic actions.  Miller
and Shanahan \cite{miller94narratives} discuss narratives in the
situation calculus (sequences of actions about which one has
incomplete information).  Reiter \cite{reiter93proving} discusses
methods to prove that certain properties are true in all states
accessible from a given state by a sequence of actions.  He discusses
formalizing databases using the situation calculus and reasoning about
their properties.  Finzi, Pirri, and Reiter \cite{finzi00open} discuss
open world planning and the associated theorem proving problem, using
essentially the formalism of Reiter \cite{Reiter:91}.  They also
convert the planning problem to a propositional problem.  Scherl and
Levesque \cite{scherl93frame} consider the frame problem in the
context of actions that gain knowledge, such as looking up a telephone
number.  They also use Reiter's formalism.  It is interesting that
they use Moore's \cite{Moore:85} ``possible world'' formalism for
knowledge, which is closely related to the modal aspect formalisms
given below.  St\"orr and Thielscher \cite{thielscher:CL00} give an
equational formulation of the situation calculus, which is based on
the work of H\"olldobler and Thielscher \cite{thielscher:AMAI95}; this
is a different formalism from that of Reiter, and also distinct from
the aspect formalism.  In their formalism, the {\em fluent calculus},
situations are represented essentially as logical formulae, that is,
conjunctions of fluents that are true in the situation.

Next a variety of formalisms for justifying the non-interference axiom
for aspects are presented.  These formalisms will aid in understanding
this axiom, as well as providing guidance about when and how this
axiom may be used.

\section{Relational Formalisms}

\subsection{The Relational Formalism (Existential Version)}
In this formalism, there are {\em relations} between situations,
denoted by $R_{\alpha}$ where $\alpha$ is an aspect.  The intuitive
meaning of $R_{\alpha}(s,t)$ is that the situation $t$ is an
$\alpha$-aspect of situation $s$.  Thus if $s$ represents a building,
$t$ can represent one room of the building.
This formalism has the
following axioms:
\begin{equation}
\forall a \alpha \beta(d(\alpha,\beta) \wedge a:\beta \supset
(R_{\alpha}(s,t) \equiv R_{\alpha}(a(s),t))) \label{relation.1}
\end{equation}
\begin{equation}
\forall p \alpha (p : \alpha \supset \exists q \forall s (p(s) \equiv
\exists t (q(t) \wedge R_{\alpha}(s,t)))) \label{relation.2}
\end{equation}
In words, axiom \ref{relation.1} says that $\alpha$ and $\beta$ are
non-interfering if the $\alpha$-aspect of situation $s$ is the same as
the $\alpha$-aspect of situation $a(s)$ for an action $a$ of aspect
$\beta$.  For example, painting one room of a building does not affect
the other rooms of the building.  Axiom \ref{relation.2} states that a
fluent $p$ has aspect $\alpha$ if $p(s)$ only depends on the $\alpha$
aspect of situation $s$.  The $q$ in axiom \ref{relation.2} is denoted
by $p_{\alpha}$.
For example, let $p(s)$ be the property that a building is heated in
situation $s$.  If the heater is in room $r_4$, and $q(t)$ specifies
that the heater is on, and $\alpha$ is $r_4$,
then $p(s) \equiv \exists t (q(t) \wedge
R_{\alpha}(s,t))$.  Intuitively, the building is heated if the heater
is on, so that an action that does not influence room $r_4$ will not
affect fluent $p$.  To be more precise, because this version is
existential, there can be several heaters;  $R_{\alpha}(s,t)$ means
that $t$ is {\em one} of the heaters of the building, and the building
is heated if any one of the heaters is on.
Also, if the circulation of blood through the body is
a property of the heart, then an action that does not influence the heart
will not affect the circulation.  In these examples, the predicate $q$ has
a natural interpretation in the problem, but the formalism can be used
even when $q$ does not have a natural interpretation.
\begin{theorem}
\label{exist.relation.theorem.1}
The non-interference axiom, axiom \ref{intro.2}, follows from axioms
\ref{relation.1} and \ref{relation.2}.
\end{theorem}
\begin{proof}
Suppose $p : \alpha$ and $a : \beta$ and $d(\alpha, \beta)$.
From $p : \alpha$ and formula \ref{relation.2},
\begin{equation}
\exists q \forall s (p(s) \equiv \exists t(q(t) \wedge R_{\alpha}(s,t))).
\end{equation}
From $a : \beta$, $d(\alpha, \beta)$, and formula \ref{relation.1},
\begin{equation}
R_{\alpha}(s,t) \equiv R_{\alpha}(a(s),t))
\end{equation}
Therefore $p(s) \equiv \exists t(q(t) \wedge R_{\alpha}(s,t)))
\equiv \exists t(q(t) \wedge R_{\alpha}(a(s),t))) \equiv p(a(s))$.
\end{proof}

\subsection{The Relational Formalism (Universal Version)}
This is the same as the existential version of the relational formalism
except that axiom \ref{relation.2} has a universal quantifier instead of
an existential quantifier:
\begin{equation}
\forall a \alpha \beta(d(\alpha,\beta) \wedge a:\beta \supset
(R_{\alpha}(s,t) \equiv R_{\alpha}(a(s),t)))
\label{relation.3}
\end{equation}
\begin{equation}
\forall p \alpha (p : \alpha \supset \exists q \forall s (p(s) \equiv
\forall t (R_{\alpha}(s,t)\supset q(t)))) \label{relation.4}
\end{equation}
As an illustration,
because this version is
universal, there can be several heaters;  $R_{\alpha}(s,t)$ means
that $t$ is {\em one} of the heaters of the building, but the building
is only heated ($p(s)$) 
if all of the heaters are on ($\forall t (R_{\alpha}(s,t)\supset q(t))$,
where $q(t)$ means that heater $t$ is on).
\begin{theorem}
\label{exist.relation.theorem.2}
Axiom \ref{intro.2} follows from axioms \ref{relation.3} and
\ref{relation.4}.
\end{theorem}
\begin{proof}
Suppose $p : \alpha$ and $a : \beta$ and $d(\alpha, \beta)$.
From $p : \alpha$ and formula \ref{relation.4},
\begin{equation}
\exists q \forall s (p(s) \equiv \forall t(R_{\alpha}(s,t)\supset q(t))).
\end{equation}
From $a : \beta$, $d(\alpha, \beta)$, and formula \ref{relation.3},
\begin{equation}
R_{\alpha}(s,t) \equiv R_{\alpha}(a(s),t)
\end{equation}
Therefore $p(s) \equiv \forall t(R_{\alpha}(s,t)\supset q(t))
\equiv \forall t(R_{\alpha}(a(s),t)\supset q(t)) \equiv p(a(s))$.
\end{proof}

\subsection{The Sequential Relational Formalism (Existential Version)}
For this formalism, $R_{\alpha_1, \dots, \alpha_n}(s,t)$ is defined as
$\exists u(R_{\alpha_1}(s,u) \wedge R_{\alpha_2, \dots, \alpha_n}(u,t)$
for $n > 1$.
This formalism has the
following axioms:
\begin{equation}
\forall a \vv{\alpha} \vv{\beta}(
d(\vv{\alpha},\vv{\beta}) \wedge a:\vv{\beta} \supset
(R_{\vv{\alpha}}(s,t) \equiv R_{\vv{\alpha}}(a(s),t))) \label{seq.relation.1}
\end{equation}
\begin{equation}
\forall p \vv{\alpha} (p : \vv{\alpha} \supset \exists q \forall s (p(s) \equiv
\exists t (q(t) \wedge R_{\vv{\alpha}}(s,t)))) \label{seq.relation.2}
\end{equation}
The $q$ in axiom \ref{seq.relation.2} is denoted
by $p_{\vv{\alpha}}$.
\begin{theorem}
Axiom \ref{seq.intro.2} follows from axioms \ref{seq.relation.1} and
\ref{seq.relation.2}.
\end{theorem}
\begin{proof}
Similar to the proof of theorem \ref{exist.relation.theorem.1}
\end{proof}

This formalism has some additional properties relating sequences of
aspects:
\begin{equation}
\forall p (p : \alpha_1 \dots \alpha_n \supset p: \alpha_1 \dots \alpha_{n-1})
\label{seq.relation.extra.1}
\end{equation}
Intuitively, a fluent is a property of a state if it is a property of a
city in the state.
\begin{equation}
0 \leq k < m \wedge
d(\alpha_1 \dots \alpha_k, \beta_1 \dots \beta_n) \supset
d(\alpha_1 \dots \alpha_m, \beta_1 \dots \beta_n)
\label{seq.relation.extra.2}
\end{equation}
Intuitively, an action that does not influence a state will not
influence any city in the state.
Also, if $d(\vv{\alpha},\vv{\beta})$ is defined by
$\exists i(\alpha_i \neq \beta_i)$ then
in addition
\begin{equation}
0 \leq k < n \wedge
d(\alpha_1 \dots \alpha_m, \beta_1 \dots \beta_k) \supset
d(\alpha_1 \dots \alpha_m, \beta_1 \dots \beta_n)
\label{seq.relation.extra.3}
\end{equation}
Intuitively, if an action that only influences a state does not influence
city X, then an action that only influences a city in the state, does
not influence city X.
The proofs of these properties (\ref{seq.relation.extra.1},
\ref{seq.relation.extra.2},
\ref{seq.relation.extra.3})
are straightforward.

\subsection{The Sequential Relational Formalism (Universal Version)}

This is the same as the existential version of the sequential
relational formalism except that axiom \ref{seq.relation.2} has a
universal quantifier instead of an existential quantifier:
\begin{equation}
\forall a \vv{\alpha} \vv{\beta}(
d(\vv{\alpha},\vv{\beta}) \wedge a:\vv{\beta} \supset
(R_{\vv{\alpha}}(s,t) \equiv R_{\vv{\alpha}}(a(s),t)))
\label{seq.relation.3}
\end{equation}
\begin{equation}
\forall p \vv{\alpha} (p : \vv{\alpha} \supset \exists q \forall s (p(s) \equiv
\forall t (R_{\vv{\alpha}}(s,t)\supset q(t)))) \label{seq.relation.4}
\end{equation}
Also, $R_{\alpha_1, \dots, \alpha_n}(s,t)$ is defined as above.
\begin{theorem}
Axiom \ref{seq.intro.2} follows from axioms \ref{seq.relation.3} and
\ref{seq.relation.4}.
\end{theorem}
\begin{proof}
Similar to the proof of theorem \ref{exist.relation.theorem.2}.
\end{proof}
The properties \ref{seq.relation.extra.1} and
\ref{seq.relation.extra.2}
hold for this formalism as well, as does
\ref{seq.relation.extra.3} under the appropriate definition of $d$.

\section{Functional Formalisms}
\subsection{Simple Functional Formalism}
If in the relational formalism, all $R_{\alpha}$ are in fact
functions, that is, for all $s$ there is exactly one $t$ such that
$R_{\alpha}(s,t)$, then a functional formalism can be used.  In
this case, $t$ is a function of $s$, denoted $f_{\alpha}(s)$.
Intuitively, if the four rooms of a building correspond to four
distinct aspects, then each aspect defines a function of the building.
If an aspect of a class is an arbitrary student in the class,
then this is not a function, so one cannot use a functional formalism.
In a straightforward manner, one obtains the following axioms for the
functional formalism.
\begin{equation}
\forall a \alpha \beta(d(\alpha,\beta) \wedge a:\beta \supset f_{\alpha}(s) =
f_{\alpha}(a(s)))
\label{function.1}
\end{equation}
\begin{equation}
\forall p \alpha (p : \alpha \supset \exists q \forall s (p(s) \equiv
q(f_\alpha(s))))
\label{function.2}
\end{equation}
As an illustration,
because this version is
functional, there is just one heater;  $f_{\alpha}(s)$ is the
heater of building $s$, and
the building is heated
($p(s)$) 
if the heater is on ($\forall s (p(s) \equiv
q(f_\alpha(s)))$
where $q(t)$ means that heater $t$ is on).
\begin{theorem}
\label{simple.function.1}
Axiom \ref{intro.2} follows from axioms \ref{function.1} and
\ref{function.2}.
\end{theorem}
\begin{proof}
Suppose $p : \alpha$ and $a : \beta$ and $d(\alpha, \beta)$.
From $p : \alpha$ and formula \ref{function.2},
\begin{equation}
\exists q \forall s (p(s) \equiv q(f_{\alpha}(s))).
\end{equation}
From $a : \beta$, $d(\alpha, \beta)$, and formula \ref{function.1},
\begin{equation}
f_{\alpha}(s) = f_{\alpha}(a(s)).
\end{equation}
Therefore $p(s) \equiv q(f_{\alpha}(s)) \equiv q(f_{\alpha}(a(s)))
\equiv p(a(s))$.
\end{proof}

\subsection{Sequential Functional Formalism}
This corresponds to the sequential relational formalism when each
relation $R_{\alpha}$ is a function, as above.  One then has
\begin{equation}
f_{\alpha_1 \dots \alpha_n} = f_{\alpha_n}f_{\alpha_{n-1}} \dots
f_{\alpha_2}f_{\alpha_1}
\end{equation}
\begin{equation}
\forall a \vv{\alpha} \vv{\beta} (
d(\vv{\alpha},\vv{\beta}) \wedge a:\vv{\beta} \supset f_{\vv{\alpha}}(s) =
f_{\vv{\alpha}}(a(s)))
\label{seq.function.1}
\end{equation}
\begin{equation}
\forall p \vv{\alpha} (p : \vv{\alpha} \supset \exists q \forall s (p(s) \equiv
q(f_\vv{\alpha}(s))))
\label{seq.function.2}
\end{equation}
\begin{theorem}
Axiom \ref{seq.intro.2} follows from axioms \ref{seq.function.1} and
\ref{seq.function.2}.
\end{theorem}
\begin{proof}
Similar to the proof of theorem \ref{simple.function.1}
\end{proof}

The sequential functional formalism also has properties
\ref{seq.relation.extra.1} and \ref{seq.relation.extra.2},
and \ref{seq.relation.extra.3} holds if $d$ is defined as stated.

\section{Collective Formalisms}
The preceding formalisms help to justify the non-interference axiom.
However, they do not give insight into the definition of $d$.  It
would be helpful to derive the axiomatization of $d$ as well from more
basic properties, to help to understand this predicate and to
provide guidance about how to axiomatize it.  When $d(\alpha,\beta)$
is defined as $\alpha \neq \beta$, the preceding discussion is
sufficient, because it seems reasonable that actions and fluents that
influence different aspects of a situation will not interact.  However,
if $d(\alpha,\beta)$ is defined by $\alpha \not\in \beta$ or
$\alpha \cap \beta = \phi$, then more justification is appropriate.
What is the meaning of the underlying elements of these sets, in a
general context, and what is the meaning of this particular definition
of $d$?  The following formalisms help to answer these questions.

\subsection{The Collective Relational Formalism (Existential Version)}
In this formalism, as before, there are relations between situations.
Aspects of actions and fluents
are assumed to be sets of elements of some underlying
set ${\cal A}$ (for example, the students in a class, or the cities in
a state).  For {\em each} $x \in {\cal A}$, there is a relation
$R_x$ on situations.
The intuitive
meaning of $R_x(s,t)$ is that the situation $t$ is the
$x$-aspect of situation $s$.  Thus if $s$ represents a class, $x$ can
be an aspect corresponding to a student in the class, and
$t$ is a situation representing the properties of that student.
Also, $d(\alpha,\beta)$ iff $\alpha \cap \beta = \phi$.
This formalism has the
following axioms:
\begin{equation}
\forall a x \beta(x \not\in \beta \wedge a:\beta \supset
(R_x(s,t) \equiv R_x(a(s),t))) \label{crelation.1}
\end{equation}
\begin{equation}
\forall p \alpha (p : \alpha \supset \exists q \forall s (p(s) \equiv
(\forall x \in \alpha) \exists t (q_x(t) \wedge R_x(s,t)))) \label{crelation.2}
\end{equation}
In words, axiom \ref{crelation.1} says that action $a$ of aspect $\beta$
can only influence the $x$ portions of situation $s$ for
$x \in \beta$.  Thus
if $x \not\in \beta$ then
the $x$ portion of situation $s$ is the same as
the $x$ portion of situation $a(s)$.
For example, if an action has aspect $\{r_1,r_2\}$ where $r_i$ are
rooms of a building, then the action can only influence these rooms,
so it does not influence room $r_3$ because $r_3 \not\in \{r_1,r_2\}$.
Axiom \ref{crelation.2} states that a
fluent $p$ has aspect $\alpha$ if $p(s)$ only depends on the $x$ aspects
of situation $s$ for $x \in \alpha$.
That is, there is a predicate $q$ such that $p(s)$ iff for all $x$
in $\alpha$, there is some situation
$t$ that is an $x$ aspect of $s$ such that $q_x(t)$.
As an example, consider a university $s$ to be superior ($p(s)$)if
every faculty member $x$ of $s$ is above average in performance
($q_x(t)$).
Then $p : \alpha$ where $\alpha$ is the set of faculty members of
the university.  Let $a$ be the action of increasing the student
enrollment.  Then $a : \beta$ where $\beta$ is the set of students
of the university.  Axiom \ref{crelation.1} states that action
$a$ does not influence any of the faculty members of the university.
Axiom \ref{crelation.2} states that the fluent $p$ only depends on
the faculty members; $R_x(s,t)$ if $t$ is the situation representing
faculty member $x$ and $q_x(t)$ if $t$ is a superior faculty member.
From these axioms it follows that the university will still be superior
if student enrollment increases.
\begin{theorem}
\label{exist.crelation.theorem.1}
The non-interference axiom, axiom \ref{intro.2},
with $d$ defined by $\alpha \cap \beta \neq \phi \supset
d(\alpha,\beta)$, follows from axioms
\ref{crelation.1} and \ref{crelation.2}.
\end{theorem}
\begin{proof}
Suppose $p : \alpha$ and $a : \beta$ and $\alpha \cap \beta \neq \phi$.
From $p : \alpha$ and formula \ref{crelation.2},
\begin{equation}
\exists q \forall s (p(s) \equiv
(\forall x \in \alpha) \exists t (q_x(t) \wedge R_x(s,t)))
\end{equation}
From $a : \beta$ and formula \ref{crelation.1},
\begin{equation}
\forall x (x \not\in \beta \supset
(R_x(s,t) \equiv R_x(a(s),t)))
\end{equation}
Therefore $p(s) \equiv (\forall x \in \alpha)
\exists t(q_x(t) \wedge R_x(s,t))
\equiv (\forall x \in \alpha)
\exists t(q_x(t) \wedge R_x(a(s),t)) \equiv p(a(s))$.
\end{proof}

\subsection{The Collective Relational Formalism (Universal Version)}
This is the same as the existential version of the relational formalism
except that axiom \ref{crelation.2} has a universal quantifier instead of
an existential quantifier:
\begin{equation}
\forall a x \beta(x \not\in \beta \wedge a:\beta \supset
(R_x(s,t) \equiv R_x(a(s),t))) \label{crelation.3}
\end{equation}
\begin{equation}
\forall p \alpha (p : \alpha \supset \exists q \forall s (p(s) \equiv
(\forall x \in \alpha) \forall t (R_x(s,t) \supset q_x(t)))) \label{crelation.4}
\end{equation}
This has the same intuition as the existential version, except that
each student and faculty member corresponds to a set of situtations,
all of which must have the property $q_x$.
\begin{theorem}
\label{exist.crelation.theorem.2}
The non-interference axiom, axiom \ref{intro.2},
with $d$ defined by $\alpha \cap \beta \neq \phi \supset
d(\alpha,\beta)$, follows from axioms
\ref{crelation.3} and \ref{crelation.4}.
\end{theorem}
\begin{proof}
Suppose $p : \alpha$ and $a : \beta$ and $\alpha \cap \beta \neq \phi$.
From $p : \alpha$ and formula \ref{crelation.4},
\begin{equation}
\exists q \forall s (p(s) \equiv
(\forall x \in \alpha) \forall t (R_x(s,t) \supset q_x(t)))
\end{equation}
From $a : \beta$ and formula \ref{crelation.3},
\begin{equation}
\forall x (x \not\in \beta \supset
(R_x(s,t) \equiv R_x(a(s),t)))
\end{equation}
Therefore $p(s) \equiv (\forall x \in \alpha)
\forall t(R_x(s,t) \supset q_x(t))
\equiv (\forall x \in \alpha)
\exists t(R_x(a(s),t) \supset q_x(t)) \equiv p(a(s))$.
\end{proof}

\subsection{The Collective Functional Formalism}
This is analogous to the functional formalism, with collective aspects:
\begin{equation}
\forall a x \beta(x \not\in \beta \wedge a:\beta \supset f_x(s) =
f_x(a(s)))
\label{cfunction.1}
\end{equation}
\begin{equation}
\forall p \alpha (p : \alpha \supset \exists q \forall s (p(s) \equiv
(\forall x \in \alpha) q_x(f_x(s))))
\label{cfunction.2}
\end{equation}
The intuition is as above, but more natural;  for every aspect $x$ there
is a unique situation $f_x(s)$ representing $x$.
\begin{theorem}
\label{simple.cfunction.1}
Axiom \ref{intro.2}
with $d$ defined by $\alpha \cap \beta \neq \phi \supset
d(\alpha,\beta)$, follows from axioms
\ref{cfunction.1} and
\ref{cfunction.2}.
\end{theorem}
\begin{proof}
Suppose $p : \alpha$ and $a : \beta$ and $\alpha \cap \beta \neq \phi$.
From $p : \alpha$ and formula \ref{cfunction.2},
\begin{equation}
\exists q \forall s (p(s) \equiv (\forall x \in \alpha) q_x(f_x(s))).
\end{equation}
From $a : \beta$ and formula \ref{cfunction.1},
\begin{equation}
x \not\in \beta \supset f_x(s) =
f_x(a(s))
\end{equation}
Therefore $p(s) \equiv (\forall x \in \alpha) q_x(f_x(s)) \equiv
(\forall x \in \alpha) q_x(f_x(a(s))) \equiv p(a(s))$.
\end{proof}

It would be possible to extend the collective formalisms to sequential
collective formalisms, as well.

\section{Modal Formalisms}
There are four more aspect-based situation calculus formalisms based
on modal logic \cite{Emerson:90}.  These formalisms make use of the
modal operators $[\;]$ and $<>$.  In modal logic, $[\;] A$ means
``necessarily $A$'' and $<> A$ means ``possibly $A$.''  The modal
situation calculus formalisms introduce a number of modal operators,
one for each action and fluent.  Thus if $a$ is an action, $[a]A$
means that after $a$ executes, assertion $A$ is necessarily true.  If
$\alpha$ is an aspect, then $[\alpha]A$ means that in all worlds
obtained from the current world by relation $R_{\alpha}$, assertion
$A$ is true.  Also, $\langle \alpha \rangle A$ means that in some
world obtained from the current world by relation $R_{\alpha}$,
assertion $A$ is true.  The frame axiom $F(a,p)$ is expressed in the
modal formalisms as the following assertion:
\begin{equation}
p \equiv [a]p
\end{equation}
Intuitively, this means that p is true in the current world if and only
if it is true in all worlds reachable by action $a$.
Modal logic has the same rules as first-order logic plus additional
ones;  the only rules we need are the following:
\begin{equation}
(X \supset Y) \supset ([a]X \supset [a]Y).
\end{equation}
\begin{equation}
(X \supset Y) \supset ([\alpha]X \supset [\alpha]Y).
\end{equation}
\begin{equation}
(X \supset Y) \supset (\langle \alpha \rangle X \supset
\langle \alpha \rangle Y).
\end{equation}
We also need the axioms
\begin{equation}
\label{aspect.preserve.1}
p:\alpha \supset [a](p:\alpha)
\end{equation}
and
\begin{equation}
a':\beta \supset [a](a':\beta)
\end{equation}
which say that aspects are preserved under actions.
Also, if a formula $X$ is provable (without assumptions), then
one can deduce the formulas $[a]X$,$[\alpha]X$, and
$\langle \alpha \rangle X$ as well.

\subsection{Simple Modal Formalism, $[\;]$ Version}
This formalism is analogous to the universal version of the relational
formaliism.  This formalism has the following axioms:
\begin{equation}
\forall a \alpha \beta (
d(\alpha,\beta) \wedge a:\beta \supset ([\alpha]X \equiv
[a][\alpha]X))
\label{simple.modal.1}
\end{equation}
\begin{equation}
\forall p \alpha (p : \alpha \supset \exists q(p \equiv [\alpha]q))
\label{simple.modal.2}
\end{equation}

\begin{theorem}
\label{simple.modal.necessary}
The non-interference axiom \ref{intro.2} follows from
formulas \ref{simple.modal.1} and \ref{simple.modal.2}.
\end{theorem}
\begin{proof}
Suppose $p : \alpha$ and $a : \beta$ and $d(\alpha, \beta)$.
From $p : \alpha$ and formula \ref{simple.modal.2},
\begin{equation}
\exists q (p \equiv [\alpha]q).
\end{equation}
From $a : \beta$, $d(\alpha, \beta)$, and formula \ref{simple.modal.1},
\begin{equation}
[\alpha]X \equiv [a][\alpha]X
\end{equation}
Therefore $p \equiv [\alpha]q \equiv [a][\alpha]q
\equiv [a]p$.
The last step $[a][\alpha]q \equiv [a]p$ makes use of $[a](p:\alpha)$
from which it is derivable.  The assertion $[a](p:\alpha)$ is
derivable from the assumption $p : \alpha$ and axiom
\ref{aspect.preserve.1}.
\end{proof}

\subsection{Simple Modal Formalism, $<>$ Version}
This formalism is analogous to the existential version of the relational
formaliism.  This formalism has the following axioms:
\begin{equation}
\forall a \alpha \beta (
d(\alpha,\beta) \wedge a:\beta \supset (\langle \alpha \rangle X
\equiv [a] \langle \alpha \rangle X))
\label{simple.modal.3}
\end{equation}
\begin{equation}
\forall p \alpha (p : \alpha \supset \exists q(p \equiv 
\langle \alpha \rangle q))
\label{simple.modal.4}
\end{equation}

\begin{theorem}
\label{simple.modal.possibly}
The non-interference axiom \ref{intro.2} follows from
formulas \ref{simple.modal.3} and \ref{simple.modal.4}.
\end{theorem}
\begin{proof}
Suppose $p : \alpha$ and $a : \beta$ and $d(\alpha, \beta)$.
From $p : \alpha$ and formula \ref{simple.modal.4},
\begin{equation}
\exists q (p \equiv \langle \alpha \rangle q).
\end{equation}
From $a : \beta$, $d(\alpha, \beta)$, and formula \ref{simple.modal.3},
\begin{equation}
\langle \alpha \rangle X \equiv [a] \langle \alpha \rangle X
\end{equation}
Therefore $p \equiv \langle \alpha \rangle q \equiv
[a] \langle \alpha \rangle q
\equiv [a]p$.  The last step makes use of the assumption $p : \alpha$ and
axiom \ref{aspect.preserve.1}, as before.
\end{proof}

\subsection{Sequential Modal Formalism, $[\;]$ Version}
This formalism is analogous to the sequential universal relational
formalism.  Sequences of aspects correspond to sequences of modal
operators.  This formalism has the following axioms:
\begin{equation}
[\alpha_1 \alpha_2 \dots \alpha_n]X \equiv
[\alpha_1][\alpha_2] \dots [\alpha_n]X
\end{equation}
\begin{equation}
\forall a \vv{\alpha}\vv{\beta}(
d(\vv{\alpha},\vv{\beta}) \wedge
a:\vv{\beta} \supset ([\vv{\alpha}]X \equiv
[a][\vv{\alpha}]X))
\label{seq.modal.1}
\end{equation}
\begin{equation}
\forall p \vv{\alpha} (p : \vv{\alpha} \supset
\exists q(p \equiv [\vv{\alpha}]q))
\label{seq.modal.2}
\end{equation}

\begin{theorem}
The non-interference axiom \ref{seq.intro.2} follows from
formulas \ref{seq.modal.1} and \ref{seq.modal.2}.
\end{theorem}
\begin{proof}
Similar to the proof of theorem \ref{simple.modal.necessary}.
\end{proof}

\subsection{Sequential Modal Formalism, $<>$ Version}
This formalism is analogous to the sequential existential
relational formalism.
This formalism has the following axioms:
\begin{equation}
\langle \alpha_1 \alpha_2 \dots \alpha_n \rangle X \equiv
\langle \alpha_1 \rangle \langle \alpha_2 \rangle \dots 
\langle \alpha_n \rangle X
\end{equation}
\begin{equation}
\forall a \vv{\alpha} \vv{\beta}(
d(\vv{\alpha},\vv{\beta}) \wedge
a:\vv{\beta} \supset (\langle \vv{\alpha} \rangle X
\equiv [a] \langle \vv{\alpha} \rangle X))
\label{seq.modal.3}
\end{equation}
\begin{equation}
\forall p \vv{\alpha} (p : \vv{\alpha} \supset \exists q(p \equiv 
\langle \vv{\alpha} \rangle q))
\label{seq.modal.4}
\end{equation}

\begin{theorem}
The non-interference axiom \ref{seq.intro.2} follows from
formulas \ref{seq.modal.3} and \ref{seq.modal.4}.
\end{theorem}
\begin{proof}
Similar to the proof of theorem \ref{simple.modal.possibly}.
\end{proof}

The sequential modal formalisms also have properties
\ref{seq.relation.extra.1} and \ref{seq.relation.extra.2},
and \ref{seq.relation.extra.3} holds as well if $d$ is defined as stated.

Collective modal formalisms could also be defined, but it seems
less natural to quantify over modalities.

\section{Constrained Formalisms}
The sequential formalisms are natural for domains that are tree structured,
but some domains have a different structure.  It is possible to represent
these as well using {\em constraints} on aspects. For example, in the
existential sequential relational formalism, the
{\em commutative} axiom
\begin{equation}
\forall s t \alpha_1 \alpha_2 (R_{\alpha_1,\alpha_2}(s,t) \equiv
R_{\alpha_2,\alpha_1}(s,t))
\label{commutative.constraint}
\end{equation}
expresses a constraint on the structure of situations.  If the set of
aspects is $\{0,1\}$, then situations without the commutative axiom have
a tree structure, but with the commutative axiom, the structure becomes
a mesh.  A mesh structure might be more appropriate for navigating in a
two-dimensional space, or finding locations on city streets.  
Of course, there are many other constraints corresponding to different
problem structures.

Such constraints require special care when specifying $d$.  Suppose
one specifies $d$ by
$\exists i(\alpha_i \neq \beta_i)
\supset d(\alpha,\beta)$.
Suppose $a:\beta$ and $p:\alpha$ where $\alpha =
\beta = (0,1)$.  Because there is no $i$ such that
$\alpha_i \neq \beta_i$, it appears that $F[a,p]$ is not derivable.
In fact, it is possible to derive $F[a,p]$ as follows:
From assertion \ref{seq.relation.1} with $\vv{\alpha} = (1,0)$ and
$\vv{\beta} = (0,1)$, one obtains
\begin{equation}
R_{(1,0)}(s,t) \equiv R_{(1,0)}(a(s),t).
\end{equation}
From $R_{(0,1)}(x,y) \equiv R_{(1,0)}(x,y)$ it follows that
\begin{equation}
R_{(0,1)}(s,t) \equiv R_{(0,1)}(a(s),t).
\label{commut.problem.1}
\end{equation}
From assertion \ref{seq.relation.2} and $p:(0,1)$ it follows that
\begin{equation}
p(s) \equiv \exists t (q(t) \wedge R_{(0,1)}(s,t)).
\end{equation}
From this and assertion
\ref{commut.problem.1} it follows that
$p(s) \equiv p(a(s))$, which is $F[a,p]$.
Thus one has $F[a,p]$ for $p$ having any of the aspects
$(0,0), (0,1), (1,0), (1,1)$, which means that action $a$ has no
effects at all, under a reasonable interpretation.
To avoid this problem, it is necessary to specify $d$ differently so that
$d(\alpha,\beta)$ {\em unless} $\alpha = \beta$ is derivable using
commutativity.  Thus in this example, $d((0,1),(1,0))$ would not be
asserted.

It is also possible to obtain the effect of the commutative axiom in
other sequential formalisms.  In the universal sequential relational
formalism, it is possible to use the same commutativity axiom.
In the sequential functional formalism, instead of the commutative
axiom one has the following:
\begin{equation}
\forall s t \alpha_1 \alpha_2 f_{\alpha_1,\alpha_2} =
f_{\alpha_2,\alpha_1}.
\end{equation}
In the sequential modal formalism, the $[\;]$ version, the corresponding
commutative axiom is
\begin{equation}
[\alpha_1 \alpha_2]X \equiv [\alpha_2 \alpha_1]X
\end{equation}
and in the
sequential modal formalism, the $\langle \rangle$ version, the corresponding
commutative axiom is
\begin{equation}
\langle \alpha_1 \alpha_2 \rangle X 
\equiv \langle \alpha_2 \alpha_1 \rangle X.
\end{equation}
Of course, for a two-dimensional space, another possibility is to
have aspects of the form $(x,y)$ where $x$ and $y$ are real numbers
giving $x$ and $y$ co-ordinates.
Then it would be appropriate to specify the $d$ relation using
geometrical constraints of some kind.

\bibliography{paper}

\end{document}